\documentclass{article}

\usepackage[preprint]{neurips_2026}

\usepackage[utf8]{inputenc}
\usepackage[T1]{fontenc}
\usepackage{hyperref}
\usepackage{url}
\usepackage{booktabs}
\usepackage{amsmath,amsfonts}
\usepackage{graphicx}
\usepackage{microtype}
\usepackage{xcolor}
\usepackage{caption}
\usepackage{wrapfig}

\title{Head Similarity: Modeling Structured Whole-Head Appearance Beyond Face Recognition}

\author{%
  Yingfeng Wang \\
  United Arab Emirates \\
  University \\
  \texttt{yingfengwang161@gmail.com} \\
  \And
  Yuxuan Xiao \\
  Nanjing University of \\
  Science and Technology \\
  \texttt{xiaoyuxuan@njust.edu.cn} \\
  \And
  Shengcai Liao \\
  United Arab Emirates \\
  University \\
  \texttt{scliao@ieee.org} \\
}

\begin{document}

\maketitle

\begin{abstract}
Many vision applications require identity consistency beyond strict biometric recognition, especially under non-frontal views or when facial cues are missing. However, conventional face recognition models enforce intra-identity invariance, collapsing appearance variations such as hairstyle or styling changes into a single representation, limiting their use in appearance-sensitive scenarios. To address this limitation, we introduce \emph{Head Similarity}, a new formulation that extends identity-centric recognition to structured whole-head similarity modeling. Our approach explicitly captures intra-identity appearance variation and enforces hierarchical similarity ordering across identity and appearance states, enabling meaningful comparison even under occlusion or rear-view conditions. We construct a large-scale benchmark from long-form videos with weakly-supervised appearance states, covering diverse poses, occlusions, and temporal changes. As a first step, we develop a simple yet effective framework that jointly models identity discrimination and appearance-sensitive similarity through hierarchical supervision and identity-aware distillation. Experiments show that conventional face recognition models fail to capture appearance-dependent similarity, while our approach demonstrates the feasibility of structured whole-head similarity modeling.
\end{abstract}

\section{Introduction}
\label{sec:intro}

\begin{figure}[t]
  \centering
  \vspace{-15pt}
  \includegraphics[width=0.9\linewidth]{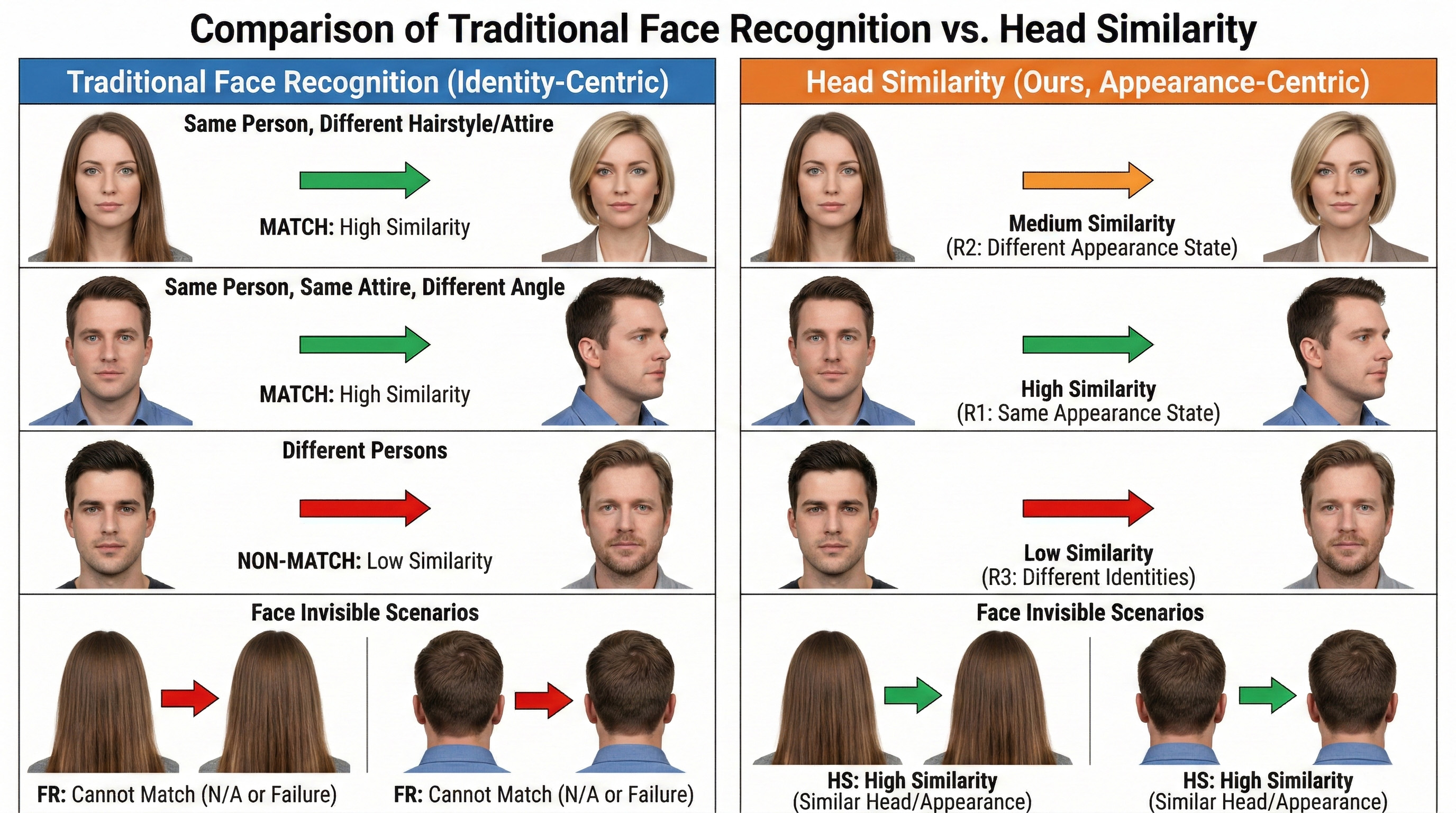}
  \vspace{-5pt}
  \caption{Conceptual comparison between identity-centric face recognition and our proposed Head Similarity. Conventional recognition collapses intra-identity appearance variations into a single compact cluster. In contrast, Head Similarity models structured appearance states within the same identity and enforces hierarchical ordering among $R_1$, $R_2$, and $R_3$.}
  \vspace{-15pt}
  \label{fig:concept}
\end{figure}

Many vision applications require \emph{identity consistency} rather than strict biometric recognition. For example, video generation, reenactment, and editing systems need a person to remain visually consistent across frames despite changes in viewpoint, expression, or hairstyle~\cite{wang2018video,song2022everybody,siarohin2019first}.
\begin{wrapfigure}{r}{0.5\linewidth}
  \centering
  \vspace{-10pt}
  \includegraphics[width=\linewidth]{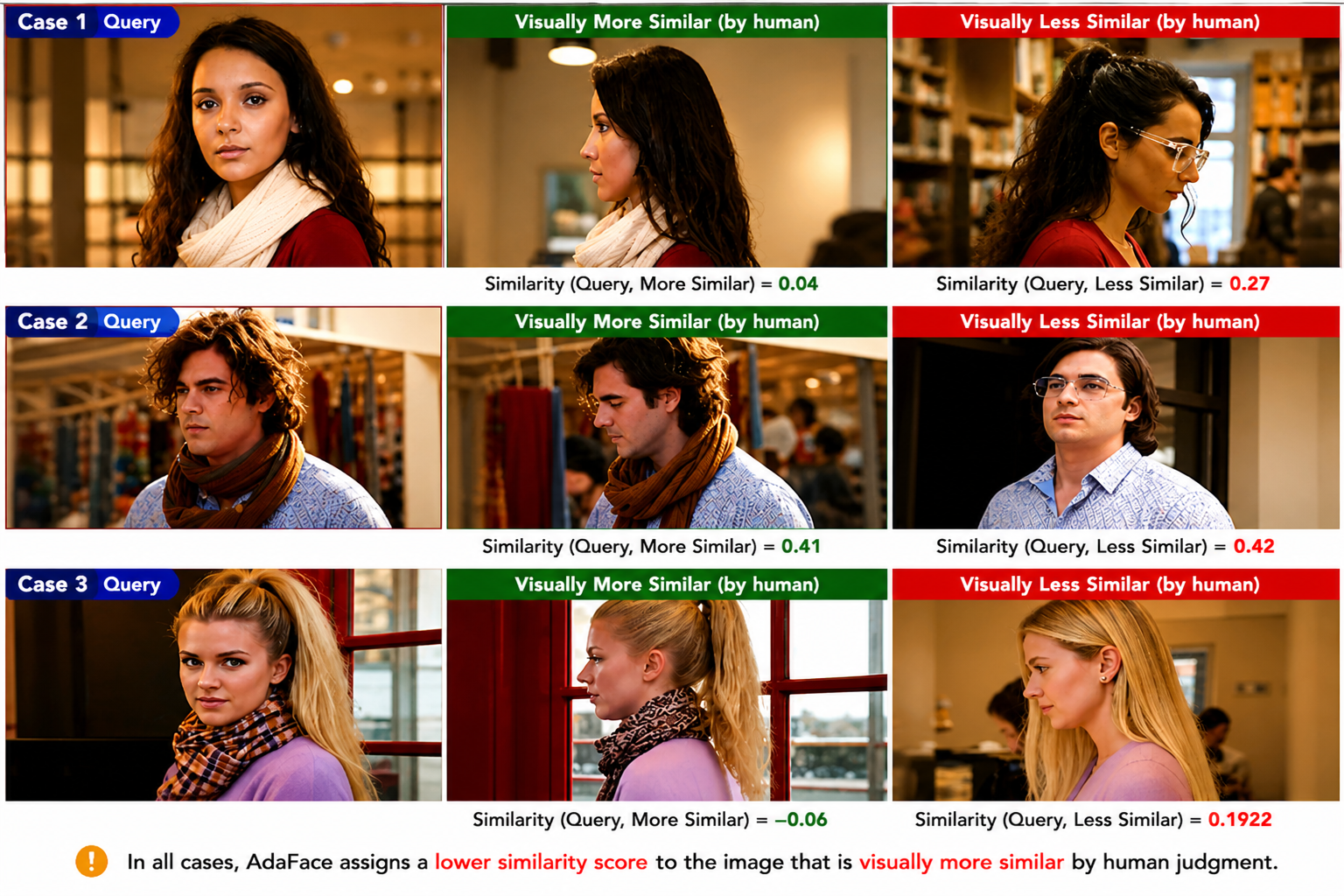}
  \vspace{-20pt}
  \caption{
  Failure cases of AdaFace on whole-head similarity.
  }
  \vspace{-10pt}
  \label{fig:adaface_failure}
\end{wrapfigure}
The goal is not to verify legal identity, but to preserve the perception that the sequence depicts the same person. Conventional face recognition is designed for a different objective. Modern systems learn identity-invariant embeddings with margin-based metric learning losses~\cite{arcface,adaface}, deliberately suppressing intra-identity variations such as hairstyle, makeup, or aging.
While effective for biometric verification, such embeddings often collapse different appearance states of the same identity into a single cluster.
As shown in Fig.~\ref{fig:adaface_failure}, AdaFace fails to preserve human-perceived appearance ordering on synthetic whole-head examples generated by Qwen~\cite{xu2025qwen3}: the visually closer sample may receive a lower similarity score.
This suggests that identity-centric embeddings are insufficient for appearance-sensitive whole-head similarity. This limitation becomes more pronounced in real-world scenarios with partial, non-frontal, or rear-view observations, where facial cues may be weak or invisible. 
In such cases, whole-head cues such as hair, contour, and accessories remain important for perceived identity consistency.
Recent hairstyle modeling and transfer works~\cite{chang2023hairnerf} further highlight the importance of head appearance beyond facial identity.
These observations reveal a gap between face similarity and holistic head-level similarity.
To address this gap, we introduce \emph{Head Similarity}, a new formulation that extends identity-centric recognition to structured whole-head appearance modeling.
Unlike conventional face recognition, which enforces intra-identity invariance, Head Similarity preserves appearance states within the same identity and enforces hierarchical similarity ordering across identity and appearance, as illustrated in Fig.~\ref{fig:concept}.
To enable systematic study, we construct a benchmark from long-form videos, where temporal grouping provides weak supervision for appearance states.
The dataset covers diverse poses, occlusions, and temporal appearance changes.
We further propose a simple Vision Transformer baseline that combines identity-aware distillation with hierarchical similarity supervision to jointly model identity preservation and appearance-sensitive similarity.
In summary, our contributions are:
\begin{itemize}
    \item[$\bullet$] We introduce \emph{Head Similarity}, a new task for structured whole-head similarity modeling beyond identity-invariant face recognition.
    \item[$\bullet$] We construct a benchmark from long-form videos with weakly supervised appearance states for evaluating hierarchical head similarity.
    \item[$\bullet$] We propose a simple baseline combining identity distillation and hierarchical similarity supervision.
    \item[$\bullet$] We show that conventional face recognition models fail to capture appearance-dependent similarity, while our method improves whole-head matching.
\end{itemize}

\section{Related Works}
\subsection{Face Recognition}
Modern face recognition learns identity-invariant embeddings with margin-based losses such as ArcFace~\cite{arcface}, CosFace~\cite{cosface}, and AdaFace~\cite{adaface}, built on ResNet~\cite{resnet} or ViT~\cite{vit}. These methods collapse intra-identity variations into compact clusters, which is effective for verification but suppresses appearance changes and degrades under extreme pose or occlusion~\cite{partialface,partialfc}. In contrast, we operate on whole-head inputs and explicitly preserve appearance variations within the same identity.

\subsection{Person Re-Identification}
Person Re-ID matches identities across cameras~\cite{pcb,mgn,transreid}, typically assuming consistent clothing. Recent works address clothing changes~\cite{gu2022clothes,qian2020long}. Unlike Re-ID, we focus on the head region and model structured appearance states within the same identity rather than enforcing invariance.

\subsection{Pose-Invariant and Occlusion-Robust Recognition}
Pose-invariant and occlusion-robust methods handle extreme views or missing regions via frontalization or feature learning~\cite{drgan,tpgan,occlusionface,partialfc}. Head360~\cite{he2024head360} extends modeling to full-head geometry. These approaches still aim at identity invariance, whereas we treat appearance variations as informative signals.

\subsection{Fine-Grained and Structured Similarity Learning}
Metric learning methods~\cite{facenet,contrastive,proxynca,supcon} structure embeddings by similarity. HIER~\cite{kim2023hier} and hyperbolic learning~\cite{gonzalez2024hyperbolic} model hierarchical relations across classes. Our work instead models hierarchical similarity within identities by leveraging video-level grouping as weak supervision.

\section{Problem Formulation}
\label{sec:problem}

\paragraph{Identity-Centric Face Recognition.}
Conventional face recognition learns identity-invariant embeddings from aligned facial crops. 
Given a dataset
\[
\mathcal{D}_{FR} = \{(x_i, u_i)\}_{i=1}^N,
\]
where $u_i$ denotes identity, the objective is to learn an embedding function $f_\theta$ such that
\[
s_\theta(x_i, x_j) > s_\theta(x_i, x_k),
\quad \text{whenever } u_i = u_j,\; u_i \neq u_k.
\]
with $s_\theta$ denoting cosine similarity. 
This formulation encourages all samples of the same identity to collapse into a single compact cluster, 
suppressing intra-identity appearance variations (e.g., hairstyle or grooming changes). 
While effective for biometric verification, it is insufficient for Head Similarity, 
which operates on whole-head images and must distinguish different appearance states of the same person.

\paragraph{Head Image Factors.}
\begin{wrapfigure}{r}{0.6\linewidth}
  \vspace{-13pt}
  \centering
  \includegraphics[width=\linewidth]{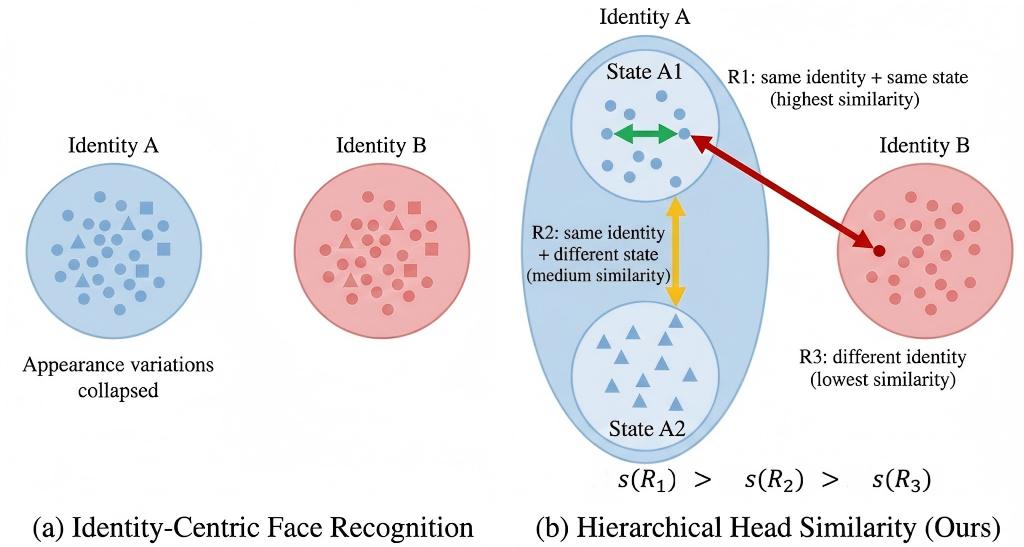}
  \vspace{-18pt}
  \caption{Illustration of the hierarchical similarity structure in the embedding space. Samples in $R_1$ (same identity, same appearance state) form the tightest clusters, samples in $R_2$ (same identity, different appearance states) lie at intermediate distances, and samples in $R_3$ (different identities) are pushed furthest apart, realizing the ordering $s(R_1) > s(R_2) > s(R_3)$.}
  \vspace{-12pt}
  \label{fig:hierarchy}
\end{wrapfigure}
We consider a head image $x$ to depend on identity $u$, appearance state $a$, and nuisance factors $e$ (illumination, background, minor pose). We define the appearance state $a$ as a configuration of identity-dependent visual attributes that remain relatively stable over a short temporal duration, including hairstyle, head shape, makeup, and accessories (e.g., glasses, earrings, or hats). In contrast, nuisance factors such as illumination, background, and minor pose variations are excluded from the definition of $a$.
Two samples are considered to share the same appearance state if these attributes are visually consistent, such that a human observer would perceive them as exhibiting the same overall head appearance despite minor variations. Traditional face recognition seeks representations dependent only on $u$, discarding $a$ and being invariant to $e$. In contrast, Head Similarity must preserve both identity and appearance state information while remaining invariant to nuisance factors. We define the dataset as
\[
\mathcal{D} = \{(x_i, u_i, a_i)\}_{i=1}^N,
\]
where $a_i$ denotes appearance state. In practice, $a_i$ is approximated using video-level grouping: frames from the same short video segment are treated as sharing the same appearance state, based on the assumption that identity-dependent attributes (e.g., hairstyle and accessories) remain stable within a short temporal window. We note that this approximation is imperfect: a single segment may contain multiple appearance states, and different segments may share the same state. Therefore, $a_i$ provides a form of weak and noisy supervision.

\paragraph{Hierarchical Similarity.}
Based on identity and appearance state, we define three pairwise relations:
\begin{align}
R_1 &= \{(i,j) \mid u_i = u_j \land a_i = a_j \}, \nonumber\\
R_2 &= \{(i,j) \mid u_i = u_j \land a_i \neq a_j \}, \nonumber\\
R_3 &= \{(i,j) \mid u_i \neq u_j \}.
\end{align}
$R_1$ corresponds to the same identity and appearance state, 
$R_2$ to the same identity but different appearance states, 
and $R_3$ to different identities.

Let $f_\theta: \mathcal{X} \rightarrow \mathbb{R}^d$ and define
\[
s_\theta(x_i, x_j) = \cos\big(f_\theta(x_i), f_\theta(x_j)\big).
\]
Head Similarity requires the hierarchical ordering
\[
s_\theta(x_i, x_j) > s_\theta(x_i, x_k) > s_\theta(x_i, x_\ell),
\quad \text{for } (i,j)\in R_1,\; (i,k)\in R_2,\; (i,\ell)\in R_3.
\]
as illustrated in Fig.~\ref{fig:hierarchy}.

Importantly, this formulation enforces that identity consistency dominates appearance similarity, 
i.e., samples of the same identity under different appearance states ($R_2$) 
should still be ranked higher than samples of different identities ($R_3$), 
even when the latter share similar appearance attributes.

\paragraph{Task Definition.}
Head Similarity thus aims to learn an embedding that 
(i) preserves identity discrimination, 
(ii) retains structured intra-identity appearance variation, and 
(iii) remains invariant to nuisance factors. 
This transforms similarity modeling from a binary identity decision problem 
into a hierarchical ranking problem over identity and appearance state.

\section{Method}
\label{sec:method}

\begin{figure*}[h]
  \vspace{-10pt}
  \centering
  \includegraphics[width=0.7\textwidth]{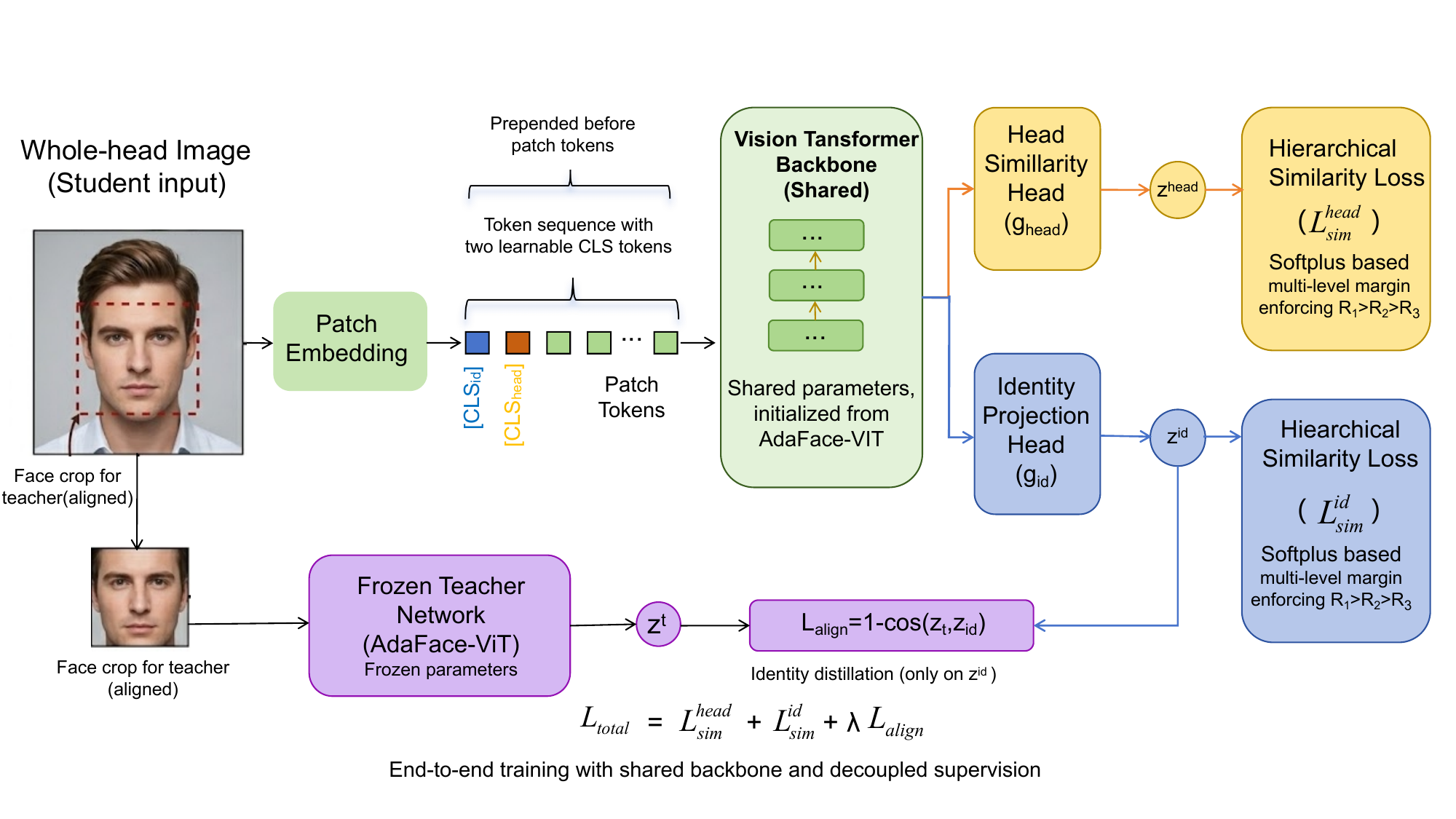}
  \vspace{-13pt}
  \caption{Overall training framework for Head Similarity. 
  A dual-CLS Vision Transformer backbone produces identity and head-similarity embeddings. 
  The identity CLS branch is aligned with a frozen face-recognition teacher via distillation, 
  while both CLS branches are supervised with hierarchical similarity loss.}
  \vspace{-15pt}
  \label{fig:framework}
\end{figure*}

As illustrated in Fig.~\ref{fig:framework}, our framework consists of three
components: a dual-CLS Vision Transformer backbone, a distillation-based
identity alignment module, and a hierarchical similarity objective. Given a
whole-head image, the backbone produces an identity embedding (identity branch) and a
head-similarity embedding (head-similarity branch). The identity branch is aligned with a frozen face-recognition teacher, while the head-similarity branch models appearance-sensitive similarity.

\subsection{Dual-CLS Vision Transformer}
\label{sec:architecture}

Building on the formulation in Sec.~\ref{sec:problem}, we aim to model structured whole-head appearance while preserving identity discrimination, and is robust to nuisance factors. Directly adapting a face-aligned backbone to whole-head inputs induces distribution shift and tends to collapse intra-identity appearance differences. We therefore adopt a dual-CLS Vision Transformer that explicitly separates identity preservation from appearance-sensitive head similarity within a shared backbone.

We instantiate the backbone with a ViT model pre-trained using AdaFace. Unlike conventional face-recognition systems that operate on aligned facial crops, our network takes whole-head images as input and can exploit cues such as hairstyle and head contour. Following prior work on multiple learnable tokens in transformers~\cite{ryoo2021tokenlearner}, we adopt a simple dual-CLS design as a baseline architecture for the proposed task.

\paragraph{Two CLS tokens.}
Given an input head image $x$, we divide it into non-overlapping patches and project them into patch tokens. We prepend two learnable tokens, $\text{CLS}^{id}$ and $\text{CLS}^{head}$, forming
\[
[\text{CLS}^{id}, \text{CLS}^{head}, x_{\text{patches}}] \in \mathbb{R}^{(2+N)\times d},
\]
where $N$ is the number of patches and $d$ is the embedding dimension. Both tokens participate in self-attention and aggregate global head-level information, but are optimized under different objectives.
\paragraph{Identity and Head-Similarity Embeddings.}
We extract two CLS tokens from the transformer and project them via $g_{id}$ and $g_{head}$, followed by L2 normalization:
\[
z^{id},\, z^{head} \in \mathbb{R}^d.
\]
The identity embedding $z^{id}$ is used at inference and trained with both distillation and hierarchical similarity supervision. The auxiliary embedding $z^{head}$ is only used during training to provide appearance-sensitive signals and is discarded at inference. Both embeddings are supervised by the hierarchical similarity objective, while $z^{id}$ is additionally aligned with a frozen teacher.

\subsection{Distillation-Based Identity Preservation}
\label{sec:distillation}

To preserve compatibility with conventional face-recognition embeddings, we align $z^{id}$ with a frozen teacher pre-trained on aligned face images. Given a face crop $x^{face}$ and a corresponding head image $x^{head}$, the teacher produces $z^t$ and the student produces $z^{id}$. 

Distillation is applied only when valid face crops are available. We minimize:
\[
\mathcal{L}_{align} = 1 - \cos(z^t, z^{id}).
\]

The teacher remains frozen. This preserves identity consistency under distribution shift from face crops to whole-head inputs, while allowing appearance variation to be modeled via the similarity objective.

\subsection{Hierarchical Similarity Learning}
\label{sec:hierarchical_loss}

Given relations $R_1, R_2, R_3$ (Sec.~\ref{sec:problem}), we construct for each anchor $x_a$: a positive $x_p \in R_1$, a semi-negative $x_{n1} \in R_2$, and a negative $x_{n2} \in R_3$. Let $z_a, z_p, z_{n1}, z_{n2}$ denote the normalized embeddings of the corresponding samples. Since embeddings are L2-normalized, cosine similarity reduces to dot product:
\[
s_{ap}=z_a^\top z_p,\quad
s_{an1}=z_a^\top z_{n1},\quad
s_{an2}=z_a^\top z_{n2}.
\]

We enforce hierarchical ordering via:
\begin{equation}
\mathcal{L}_{sim}
=
\text{Softplus}(m_1 + s_{an1} - s_{ap})
+
\text{Softplus}(m_2 + s_{an2} - s_{ap})
+
\text{Softplus}(m_3 + s_{an2} - s_{an1}).
\end{equation}
where $m_1,m_2,m_3>0$.

We further apply online hard negative mining within each batch. The loss is applied to both $z^{id}$ and $z^{head}$, jointly structuring identity and appearance-aware representations.
\subsection{Training Strategy and Dataset Construction}
\label{sec:training}

\begin{figure}[h]
  \centering
  \vspace{-10pt}
  \includegraphics[width=0.9\linewidth]{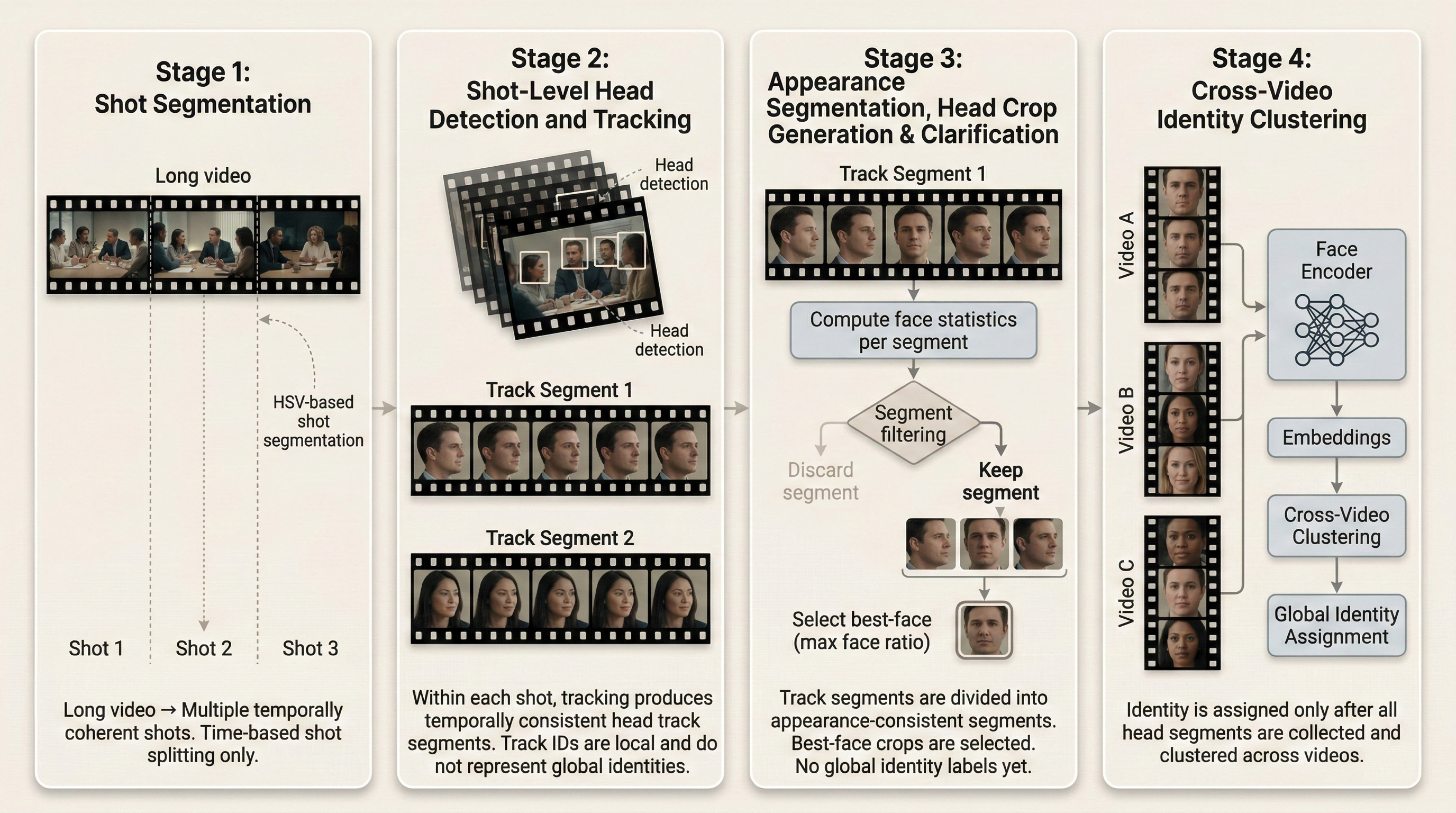}
  \vspace{-5pt}
  \caption{Pipeline of the Head Similarity dataset construction.}
  \vspace{-10pt}
  \label{fig:dataset_pipeline}
\end{figure}

We jointly leverage a head-level similarity dataset constructed in this work and large-scale face-recognition datasets for identity distillation. The pipeline (Fig.~\ref{fig:dataset_pipeline}) consists of four stages: shot segmentation, head detection and tracking, segment filtering with representative crop generation, and cross-video identity clustering.
We construct the head-level dataset through a four-stage pipeline. 
\textbf{Shot segmentation} is first performed using an HSV histogram-based method~\cite{truong2007video} to obtain temporally coherent segments. 
Within each shot, we conduct \textbf{head detection and tracking} using RetinaFace~\cite{deng2020retinaface} and short-term IoU-based tracking to produce local track segments. 
We then apply \textbf{segment filtering and representative crop selection}, retaining segments with at least 5 frames, $\geq$20\% face-visible frames, and $\geq$10\% face-invisible frames; for each segment, a \emph{best-face} frame (face area $>$50\% of the head box) is selected for identity clustering, while full head crops are preserved for training. Next, \textbf{cross-video identity clustering} is performed using face embeddings, inducing hierarchical relations where samples within a segment correspond to $R_1$, samples across segments within the same cluster correspond to $R_2$, and samples from different clusters correspond to $R_3$. For training, we combine this dataset with large-scale face-recognition data. \textbf{Face-recognition samples} from VoxCeleb2~\cite{voxceleb2} and VGGFace2~\cite{vggface2} are used for distillation, where aligned face crops are processed by a frozen teacher and corresponding head images are fed to the student. 
We adopt a \textbf{mixed-batch strategy}, where head samples provide $(R_1,R_2,R_3)$ relations with online hard negative mining, while face samples contribute only to the alignment loss. 
Finally, we apply \textbf{background randomization} by replacing backgrounds with random samples from the COCO dataset~\cite{lin2014microsoft} to reduce spurious correlations and improve robustness.

\section{Experiments}
\label{sec:experiments}
\subsection{Experimental Setup}
\label{sec:exp_setup}

\paragraph{Implementation Details.}
We use a ViT backbone initialized from an AdaFace-pretrained checkpoint. Whole-head images are resized and normalized following the teacher. We train with AdamW (lr=$1\mathrm{e}{-4}$, weight decay=$5\mathrm{e}{-2}$, batch size=256) for 30 epochs. Margins are set to $m_1=0.1$, $m_2=0.3$, and $m_3=0.2$. Experiments are conducted on two RTX 5090 GPUs.

\paragraph{Datasets.}
\begin{wraptable}{r}{0.48\linewidth}
\vspace{-15pt}
\caption{Statistics of the HeadSim-Head dataset.}
\label{tab:dataset_statistics}
\vspace{-5pt}
\centering
\fontsize{8}{8}\selectfont
\setlength{\tabcolsep}{1pt}
\begin{tabular}{lcc}
\toprule
\textbf{Metric} & \textbf{Train} & \textbf{Test} \\
\midrule
\# Identities & 731 & 187 \\
\# Videos & 6,126 & 2,489 \\
\# Images & 39,890 & 8,870 \\
Avg. states & 8 & 13 \\
\bottomrule
\end{tabular}
\vspace{-10pt}
\end{wraptable}
We use VGGFace2~\cite{vggface2} and VoxCeleb2~\cite{voxceleb2} for identity distillation, and HeadSim-Head for evaluation. \textbf{VGGFace2} is used only for distillation, where aligned face crops are processed by a frozen teacher and corresponding head images are fed to the student. \textbf{VoxCeleb2} is used for both distillation and evaluation. We follow the standard verification protocol on the official test split with strict train/test separation. \textbf{HeadSim-Head} is our benchmark dataset (Sec.~\ref{sec:training}), where samples are labeled with identity and weak appearance states. Pairs are grouped into $R_1$, $R_2$, and $R_3$ based on identity and segment relations. 

As summarized in Table~\ref{tab:dataset_statistics}, the dataset contains 731/187 identities, 6,126/2,489 videos, and 39,890/8,870 images for train/test splits, with disjoint identities. Each identity has multiple appearance states (avg. 8), covering diverse poses (38\% frontal, 53\% profile, 9\% back-view). HeadSim-Head identities do not overlap with VoxCeleb2 test identities.

\paragraph{Evaluation Metrics.}
We report verification rate (VR) at fixed false accept rates (FAR), including FAR=$10^{-2}$, $10^{-3}$, and $10^{-4}$, which are standard metrics in face verification. In addition, we report the area under the ROC curve (AUC) as a reference metric. Since AUC integrates over all operating points and may over-emphasize high-FAR regions, we primarily focus on VR at low FAR thresholds, which better reflect practical verification performance. On VoxCeleb2, VR@FAR evaluates identity verification performance under whole-head inputs. On HeadSim-Head, the same metrics measure the ability to distinguish samples across $R_1$, $R_2$, and $R_3$. Unless otherwise specified, all results use the identity embedding $z^{id}$ at inference; $z^{head}$ is used only during training.

\subsection{From Face Recognition to Whole-Head Input}
\label{sec:exp_distill}
\begin{wrapfigure}{r}{0.48\linewidth}
\vspace{-10pt}
\centering
\includegraphics[width=\linewidth]{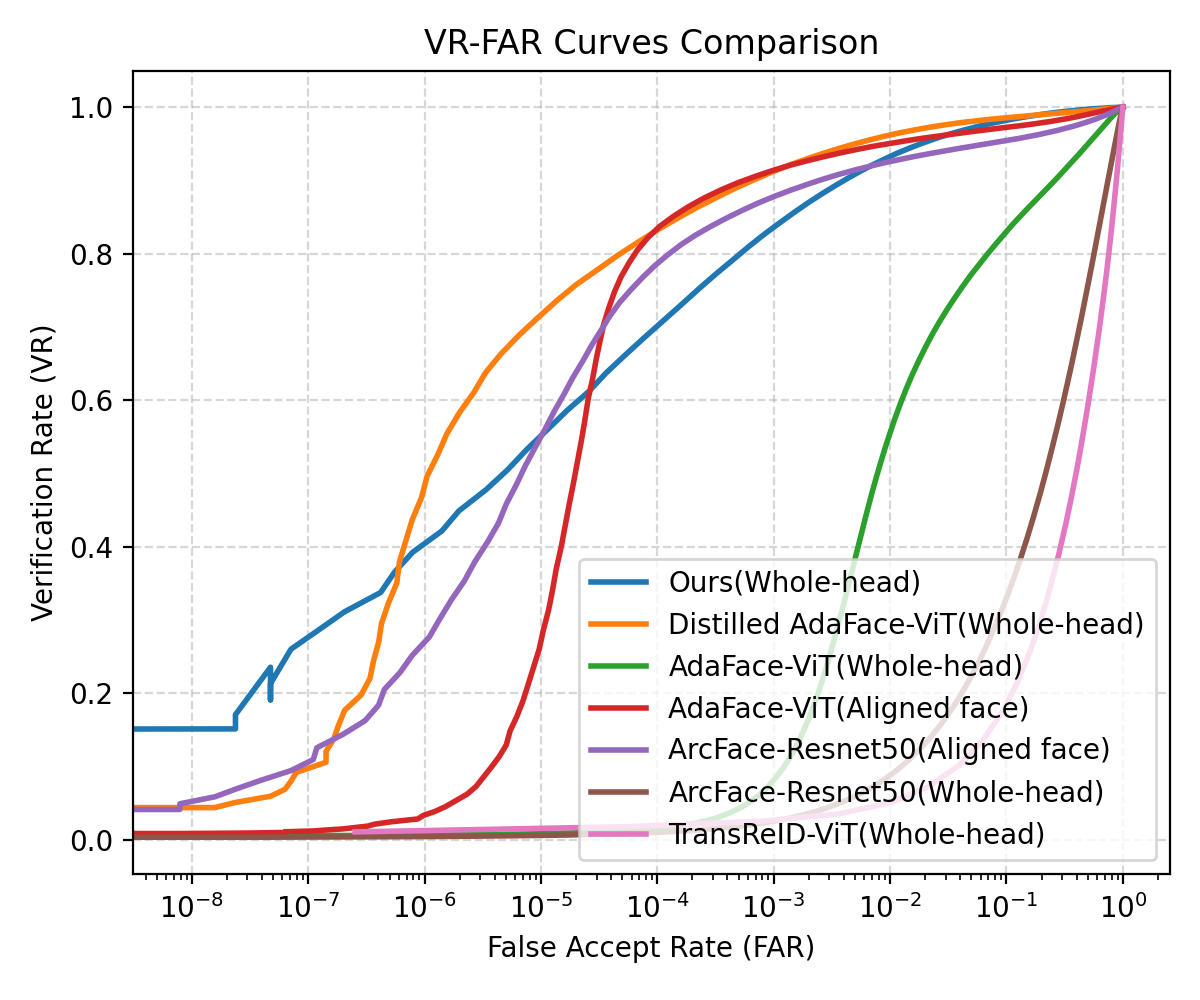}
\vspace{-20pt}
\caption{ROC curves under aligned and whole-head inputs.}
\label{fig:roc_distillation}
\vspace{-10pt}
\end{wrapfigure}
We analyze the effect of adapting face-recognition models from aligned-face inputs to unaligned whole-head images. As shown in Table~\ref{tab:identity_preservation}, directly applying face-trained models to whole-head inputs leads to a substantial performance drop, especially in the low-FAR regime. For AdaFace-ViT, VR@FAR=$10^{-3}$ decreases from 0.914 with aligned faces to only 0.078 with whole-head inputs, and ArcFace-ResNet50~\cite{arcface} shows an even larger degradation. This confirms that whole-head verification suffers from a strong distribution shift when using models trained on tightly aligned face crops.

Interestingly, although TransReID-ViT~\cite{transreid} is a strong person re-identification model, it performs poorly on identity-level verification, achieving only 0.025 VR@FAR=$10^{-3}$. This indicates that ReID models do not learn sufficiently discriminative facial identity features, as they primarily rely on global appearance cues such as clothing and body context rather than facial characteristics.

Distillation-based adaptation effectively bridges this gap: Distilled AdaFace-ViT on whole-head inputs achieves the best low-FAR performance, with VR@FAR=$10^{-3}$ of 0.912 and AUC of 0.993, closely matching the aligned-face AdaFace-ViT baseline. Our model also achieves strong whole-head verification performance, reaching an AUC of 0.993 and significantly outperforming directly applied whole-head baselines, while further supporting structured head-level similarity modeling.

\begin{table}[h]
\vspace{-10pt}
\caption{Face verification performance under aligned-face and whole-head inputs.}
\label{tab:identity_preservation}
\centering
\scriptsize
\setlength{\tabcolsep}{2pt}
\begin{tabular}{llcccc}
\toprule
Model & Input & VR@FAR=$10^{-2}$ & VR@FAR=$10^{-3}$ & VR@FAR=$10^{-4}$ & AUC \\
\midrule
AdaFace-ViT & Aligned face & 0.950 & 0.914 & 0.822 & 0.987 \\
AdaFace-ViT & Whole-head & 0.554 & 0.078 & 0.014 & 0.923 \\
Dist. AdaFace & Whole-head & 0.961 & 0.912 & 0.816 & 0.993 \\
ArcFace-Resnet50 & Aligned face & 0.925 & 0.870 & 0.784 & 0.977 \\
ArcFace-Resnet50 & Whole-head & 0.084 & 0.025 & 0.009 & 0.701 \\
TransReID-ViT & Whole-head & 0.035 & 0.025 & 0.018 & 0.570 \\
Ours & Whole-head & 0.929 & 0.826 & 0.686 & 0.993 \\
\bottomrule
\end{tabular}
\vspace{-10pt}
\end{table}

As shown in Fig.~\ref{fig:roc_distillation}, distillation-based adaptation significantly improves performance under whole-head inputs.

\subsection{Head Similarity}
\label{sec:exp_similarity}

\begin{wrapfigure}{r}{0.48\linewidth}
\vspace{-40pt}
\centering
\includegraphics[width=\linewidth]{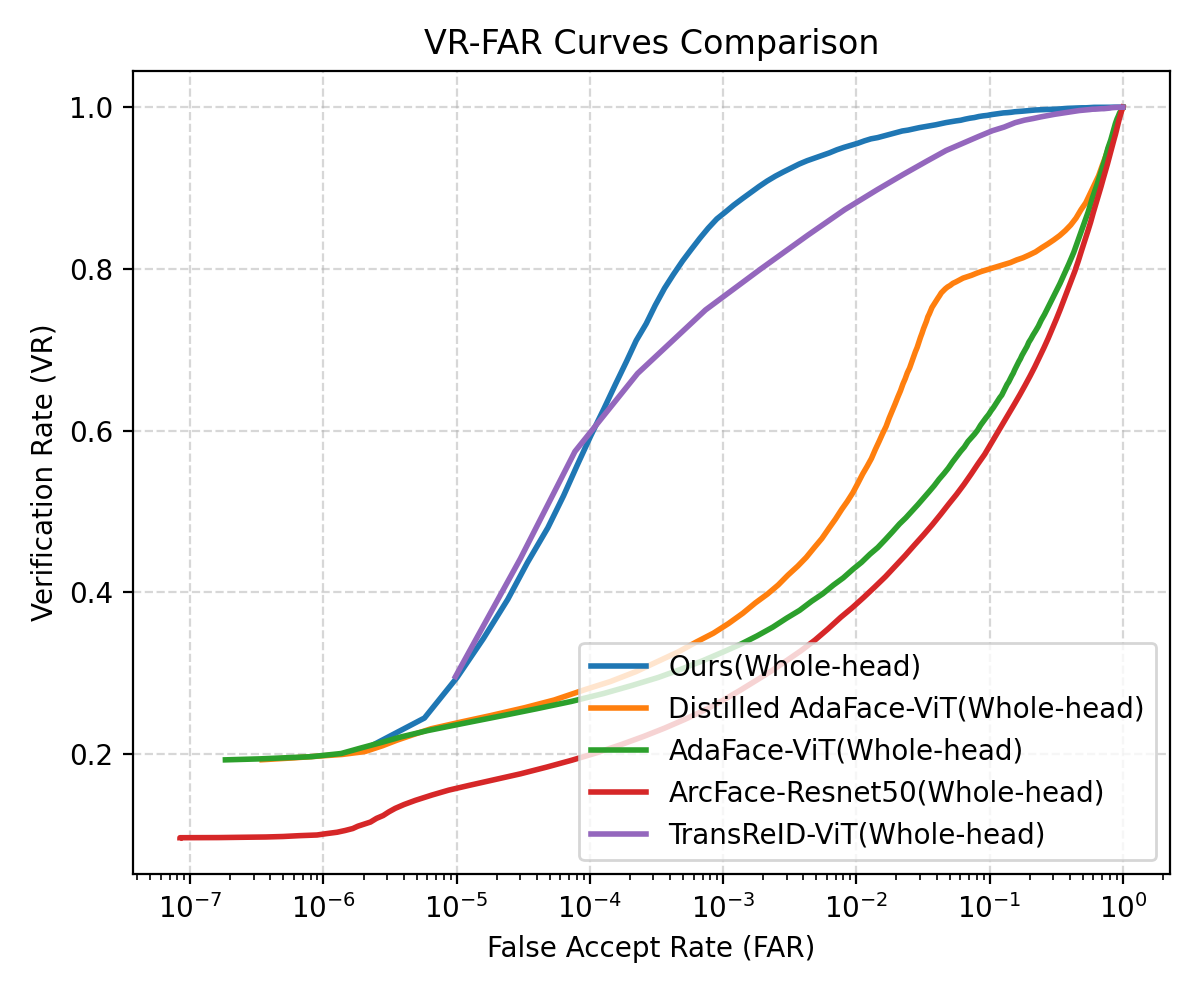}
\vspace{-20pt}
\caption{ROC curves comparison on the HeadSim-Head dataset.}
\label{fig:roc_comparison}
\vspace{-15pt}
\end{wrapfigure}

As reported in Table~\ref{tab:head_similarity_HeadSim-Head} and illustrated by the ROC curves in Fig.~\ref{fig:roc_comparison}, the vanilla AdaFace-ViT, pre-trained on aligned faces, tends to collapse intra-identity appearance variations, yielding limited performance under whole-head inputs. While identity distillation provides a moderate improvement, it remains insufficient for fine-grained appearance discrimination. TransReID-ViT achieves relatively strong performance on this task, with VR@FAR=$10^{-3}$ of 0.749 and AUC of 0.988. This is expected, as ReID models are designed to capture global appearance similarity. However, this improvement comes at the cost of identity discrimination, as shown in Table~\ref{tab:identity_preservation}, where TransReID performs poorly on identity-level verification. This indicates that ReID models rely heavily on appearance cues and do not preserve identity-sensitive features required for reliable verification.
In contrast, our framework jointly models identity discrimination and structured appearance variation. As a result, it achieves the best overall performance, with VR@FAR=$10^{-3}$ of 0.862 and AUC of 0.996, while maintaining strong identity verification capability. Notably, as shown in Fig.~\ref{fig:roc_comparison}, our model maintains high verification rates even at extremely low FAR, demonstrating its ability to distinguish subtle appearance states without sacrificing identity consistency.
\begin{table}[h]
\vspace{-15pt}
\caption{
Head similarity performance on the HeadSim-Head dataset.
}
\label{tab:head_similarity_HeadSim-Head}
\centering
\scriptsize
\setlength{\tabcolsep}{2pt}
\begin{tabular}{llcccc}
\toprule
Model & Input & VR@FAR=$10^{-2}$ & VR@FAR=$10^{-3}$ & VR@FAR=$10^{-4}$ & AUC \\
\midrule
Ours &  Whole-head & 0.953 & 0.862 & 0.558 & 0.996 \\
Dist. AdaFace-ViT & Whole-head & 0.523 & 0.349 & 0.279 & 0.876 \\
AdaFace-ViT & Whole-head & 0.428 & 0.326 & 0.265 & 0.825 \\
ArcFace-Resnet50 &  Whole-head & 0.381 & 0.266 & 0.192 & 0.798 \\
TransReID-ViT & Whole-head & 0.873 & 0.749 & 0.574 & 0.988 \\
\bottomrule
\end{tabular}
\vspace{-15pt}
\end{table}

\begin{figure*}[h]
\centering
\includegraphics[width=\textwidth]{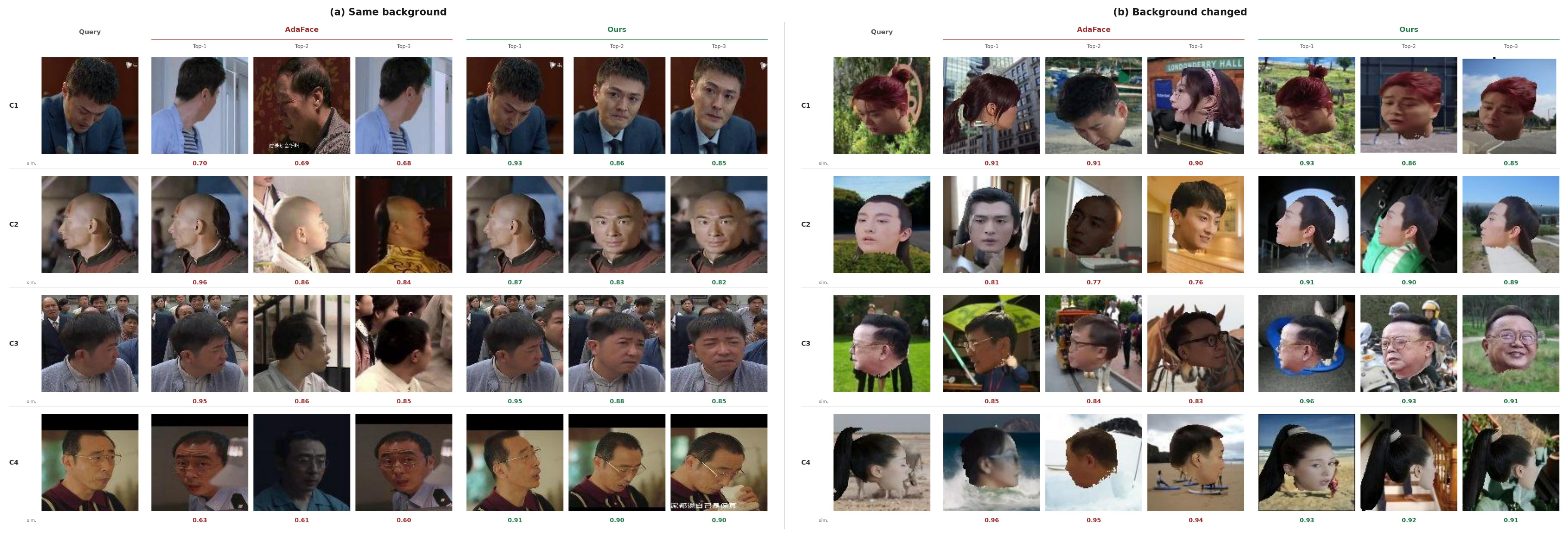}
\caption{
Top-3 retrieval results on the HeadSim-Head test set for AdaFace and our method under two settings: same background (left) and background changed (right). 
Each row shows a query and its top-3 retrieved samples with corresponding similarity scores.
}
\label{fig:retrieval_comparison}
\end{figure*}

To further analyze retrieval behavior, we visualize representative examples from the HeadSim-Head test set in Fig.~\ref{fig:retrieval_comparison}. For each query, we present the top-3 retrieved results along with similarity scores for both AdaFace and our method under two settings: same background and background changed. The background-changed setting is intentionally introduced to evaluate robustness, rather than to exploit background cues as shortcuts.
Specifically, by altering the background while preserving identity and appearance attributes, we can better isolate whether the model captures intrinsic head-level features instead of relying on contextual information. As shown in Fig.~\ref{fig:retrieval_comparison}, AdaFace often retrieves samples that are visually similar at the facial level but inconsistent in overall head appearance, particularly under background changes. In contrast, our method consistently produces identity-aligned and appearance-consistent retrievals across both settings, demonstrating robustness to background variation and stronger modeling of structured whole-head similarity.

\subsection{Task Conflict Between Identity and Appearance}
\label{sec:exp_conflict}
\begin{wrapfigure}{r}{0.50\linewidth}
\vspace{-20pt}
\centering
\includegraphics[width=\linewidth]{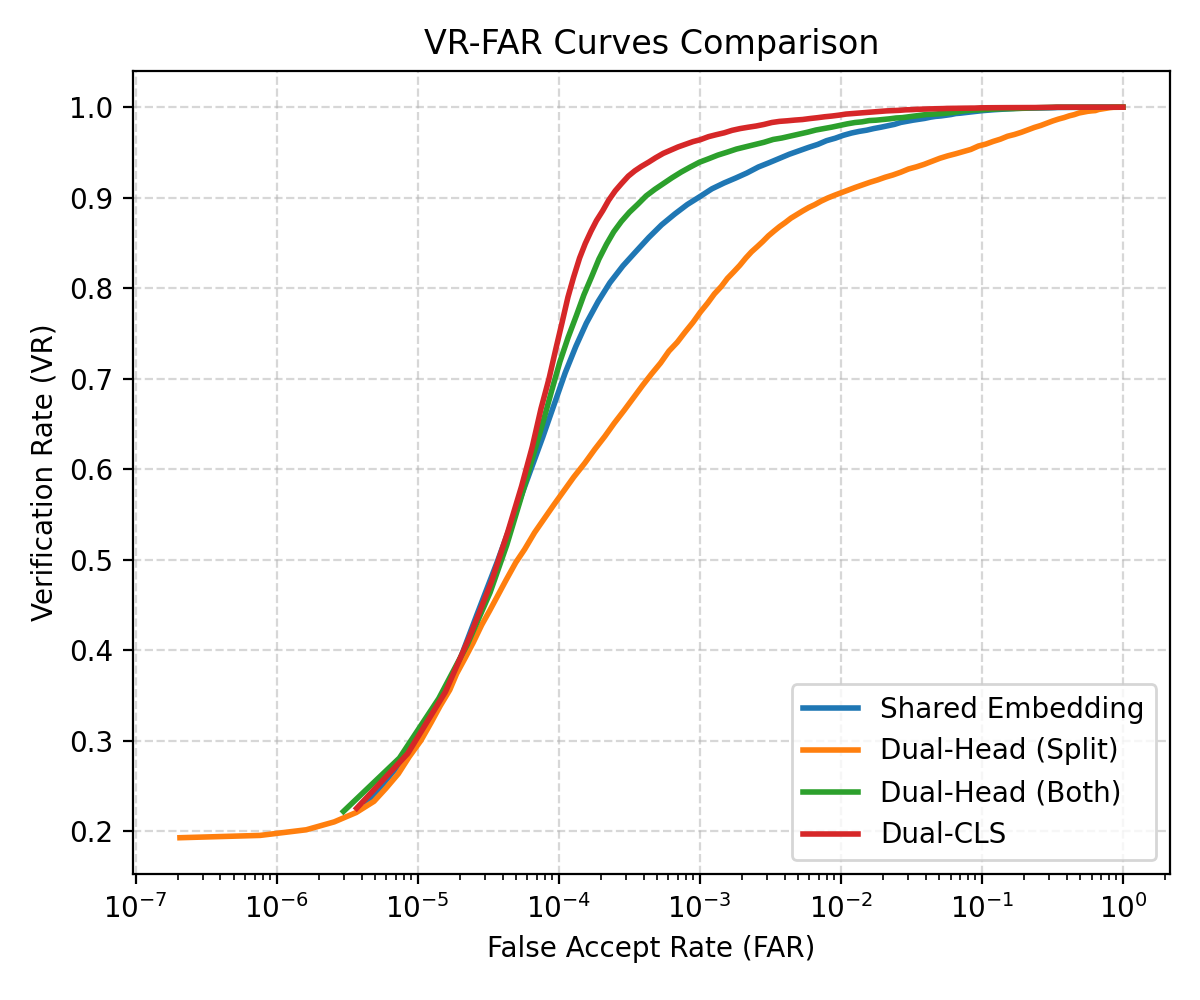}
\vspace{-20pt}
\caption{
ROC curves on HeadSim-Head for different configurations.
Dual-CLS consistently outperforms other variants.
}
\label{fig:roc_conflict}
\vspace{-10pt}
\end{wrapfigure}
To analyze the conflict between identity invariance and appearance-sensitive similarity,
we evaluate different architectural variants and loss assignments,
as summarized in Table~\ref{tab:conflict_ablation} and visualized in Fig.~\ref{fig:roc_conflict}. \textit{Shared Embedding} uses a single representation for both identity alignment and head-similarity learning, where $L_{align}$ and $L_{sim}$ are applied to the same embedding. \textit{Dual-Head (Split)} separates the two objectives into different heads, with the identity branch supervised only by $L_{align}$ and the similarity branch supervised by $L_{sim}$. \textit{Dual-Head (Both)} further applies both losses to the identity branch, while \textit{Dual-CLS} introduces two CLS tokens to decouple the two objectives at the token level. As shown in Table~\ref{tab:conflict_ablation}, the split dual-head design performs the
worst among the compared variants, with VR@FAR=$10^{-3}$ dropping to 0.762. This suggests that simply assigning the two losses to separate heads is insufficient, since the head-similarity branch lacks strong identity-preserving supervision. Applying both $L_{align}$ and $L_{sim}$ to the identity branch improves the result substantially, raising VR@FAR=$10^{-3}$ from 0.762 to 0.939.

\begin{table}[h]
\vspace{-10pt}
\caption{Ablation on loss combinations and CLS design.}
\label{tab:conflict_ablation}
\centering
\scriptsize
\setlength{\tabcolsep}{2pt}
\begin{tabular}{lcccccc}
\toprule
\textbf{Model} & \textbf{\#CLS} & \textbf{Loss on ID} & VR@FAR=$10^{-2}$ & VR@FAR=$10^{-3}$ & VR@FAR=$10^{-4}$ & AUC \\
\midrule
Shared Embedding  & 1 & $L_{align}+L_{sim}$ & 0.966 & 0.892 & 0.673 & 0.998\\
Dual-Head (Split) & 1 & $L_{align}$         & 0.903&0.762 &0.560 &0.985 \\
Dual-Head (Both)  & 1 & $L_{align}+L_{sim}$ & 0.980&0.939 &0.687 &0.999 \\
Dual-CLS (Ours)   & 2 & $L_{align}+L_{sim}$ & 0.991 &0.962 &0.734 &0.999 \\
\bottomrule
\end{tabular}
\vspace{-5pt}
\end{table}

Fig.~\ref{fig:roc_conflict} shows the same trend across FAR thresholds. The Dual-CLS
model consistently achieves the strongest ROC curve, especially in the low-FAR region.
It reaches the best VR@FAR=$10^{-2}$, VR@FAR=$10^{-3}$, and VR@FAR=$10^{-4}$ scores
of 0.991, 0.962, and 0.734, respectively. These results indicate that token-level
decoupling is more effective than post-hoc head-level separation, enabling the model
to better balance identity preservation and appearance-sensitive head similarity.

\section{Conclusion}
\label{sec:conclusion}

We introduced \emph{Head Similarity}, a formulation that extends identity-centric face recognition to structured whole-head appearance modeling. Unlike conventional methods that collapse intra-identity variations, our approach preserves appearance states while maintaining identity discrimination. To support this task, we constructed a benchmark with diverse poses, occlusions, and temporal appearance changes. We further proposed a simple framework combining identity-aware distillation with hierarchical similarity supervision. Experiments show that face recognition models fail to capture appearance-dependent similarity, while our method achieves strong whole-head matching performance without sacrificing identity verification capability.
 
\section{Ethical Considerations and Limitations}
Our work has several limitations. First, the appearance-state labels are obtained through weak temporal grouping rather than manual annotation, which may introduce noisy supervision when a video segment contains multiple appearance states or when similar states appear across different segments. Second, although our benchmark enables systematic evaluation of head similarity, its scale and diversity remain limited compared with large-scale face recognition datasets. Third, the proposed hierarchy $R_1 > R_2 > R_3$ is only a proxy for human-perceived head similarity and may not fully capture subjective perceptual judgments in ambiguous cases. Finally, our dual-CLS framework is intended as a simple baseline and does not explicitly disentangle identity, hairstyle, accessories, pose, or temporal appearance transitions.

\bibliographystyle{plainnat}
\bibliography{main}

\begin{thebibliography}{33}
\providecommand{\natexlab}[1]{#1}
\providecommand{\url}[1]{\texttt{#1}}
\expandafter\ifx\csname urlstyle\endcsname\relax
  \providecommand{\doi}[1]{doi: #1}\else
  \providecommand{\doi}{doi: \begingroup \urlstyle{rm}\Url}\fi

\bibitem[An et~al.(2021)An, Zhu, Gao, Xiao, Zhao, Feng, Wu, Qin, Zhang, Zhang, et~al.]{partialfc}
Xiang An, Xuhan Zhu, Yuan Gao, Yang Xiao, Yongle Zhao, Ziyong Feng, Lan Wu, Bin Qin, Ming Zhang, Debing Zhang, et~al.
\newblock Partial fc: Training 10 million identities on a single machine.
\newblock In \emph{Proceedings of the IEEE/CVF International Conference on Computer Vision}, pages 1445--1449, 2021.

\bibitem[Cao et~al.(2018)Cao, Shen, Xie, Parkhi, and Zisserman]{vggface2}
Qiong Cao, Li~Shen, Weidi Xie, Omkar~M Parkhi, and Andrew Zisserman.
\newblock Vggface2: A dataset for recognising faces across pose and age.
\newblock In \emph{2018 13th IEEE international conference on automatic face \& gesture recognition (FG 2018)}, pages 67--74. IEEE, 2018.

\bibitem[Chang et~al.(2023)Chang, Kim, and Kim]{chang2023hairnerf}
Seunggyu Chang, Gihoon Kim, and Hayeon Kim.
\newblock Hairnerf: Geometry-aware image synthesis for hairstyle transfer.
\newblock In \emph{Proceedings of the IEEE/CVF International Conference on Computer Vision}, pages 2448--2458, 2023.

\bibitem[Chung et~al.(2018)Chung, Nagrani, and Zisserman]{voxceleb2}
Joon~Son Chung, Arsha Nagrani, and Andrew Zisserman.
\newblock Voxceleb2: Deep speaker recognition.
\newblock \emph{arXiv preprint arXiv:1806.05622}, 2018.

\bibitem[Deng et~al.(2019)Deng, Guo, Xue, and Zafeiriou]{arcface}
Jiankang Deng, Jia Guo, Niannan Xue, and Stefanos Zafeiriou.
\newblock Arcface: Additive angular margin loss for deep face recognition.
\newblock In \emph{Proceedings of the IEEE/CVF conference on computer vision and pattern recognition}, pages 4690--4699, 2019.

\bibitem[Deng et~al.(2020)Deng, Guo, Ververas, Kotsia, and Zafeiriou]{deng2020retinaface}
Jiankang Deng, Jia Guo, Evangelos Ververas, Irene Kotsia, and Stefanos Zafeiriou.
\newblock Retinaface: Single-shot multi-level face localisation in the wild.
\newblock In \emph{Proceedings of the IEEE/CVF conference on computer vision and pattern recognition}, pages 5203--5212, 2020.

\bibitem[Dosovitskiy et~al.(2020)Dosovitskiy, Beyer, Kolesnikov, Weissenborn, Zhai, Unterthiner, Dehghani, Minderer, Heigold, Gelly, et~al.]{vit}
Alexey Dosovitskiy, Lucas Beyer, Alexander Kolesnikov, Dirk Weissenborn, Xiaohua Zhai, Thomas Unterthiner, Mostafa Dehghani, Matthias Minderer, Georg Heigold, Sylvain Gelly, et~al.
\newblock An image is worth 16x16 words: Transformers for image recognition at scale.
\newblock \emph{arXiv preprint arXiv:2010.11929}, 2020.

\bibitem[Gonzalez-Jimenez et~al.(2024)Gonzalez-Jimenez, Lionetti, Bazazian, Gottfrois, Gr{\"o}ger, Navarini, and Pouly]{gonzalez2024hyperbolic}
Alvaro Gonzalez-Jimenez, Simone Lionetti, Dena Bazazian, Philippe Gottfrois, Fabian Gr{\"o}ger, Alexander Navarini, and Marc Pouly.
\newblock Hyperbolic metric learning for visual outlier detection.
\newblock In \emph{European Conference on Computer Vision}, pages 327--344. Springer, 2024.

\bibitem[Gu et~al.(2022)Gu, Chang, Ma, Bai, Shan, and Chen]{gu2022clothes}
Xinqian Gu, Hong Chang, Bingpeng Ma, Shutao Bai, Shiguang Shan, and Xilin Chen.
\newblock Clothes-changing person re-identification with rgb modality only.
\newblock In \emph{Proceedings of the IEEE/CVF conference on computer vision and pattern recognition}, pages 1060--1069, 2022.

\bibitem[Hadsell et~al.(2006)Hadsell, Chopra, and LeCun]{contrastive}
Raia Hadsell, Sumit Chopra, and Yann LeCun.
\newblock Dimensionality reduction by learning an invariant mapping.
\newblock In \emph{2006 IEEE computer society conference on computer vision and pattern recognition (CVPR'06)}, volume~2, pages 1735--1742. IEEE, 2006.

\bibitem[He et~al.(2016)He, Zhang, Ren, and Sun]{resnet}
Kaiming He, Xiangyu Zhang, Shaoqing Ren, and Jian Sun.
\newblock Deep residual learning for image recognition.
\newblock In \emph{Proceedings of the IEEE conference on computer vision and pattern recognition}, pages 770--778, 2016.

\bibitem[He et~al.(2021)He, Luo, Wang, Wang, Li, and Jiang]{transreid}
Shuting He, Hao Luo, Pichao Wang, Fan Wang, Hao Li, and Wei Jiang.
\newblock Transreid: Transformer-based object re-identification.
\newblock In \emph{Proceedings of the IEEE/CVF international conference on computer vision}, pages 15013--15022, 2021.

\bibitem[He et~al.(2024)He, Zhuang, Wang, Yao, Zhu, Li, Zhang, Cao, and Zhu]{he2024head360}
Yuxiao He, Yiyu Zhuang, Yanwen Wang, Yao Yao, Siyu Zhu, Xiaoyu Li, Qi~Zhang, Xun Cao, and Hao Zhu.
\newblock Head360: Learning a parametric 3d full-head for free-view synthesis in 360$^\circ$.
\newblock In \emph{European Conference on Computer Vision}, pages 254--272. Springer, 2024.

\bibitem[Huang et~al.(2017)Huang, Zhang, Li, and He]{tpgan}
Rui Huang, Shu Zhang, Tianyu Li, and Ran He.
\newblock Beyond face rotation: Global and local perception gan for photorealistic and identity preserving frontal view synthesis.
\newblock In \emph{Proceedings of the IEEE international conference on computer vision}, pages 2439--2448, 2017.

\bibitem[Khosla et~al.(2020)Khosla, Teterwak, Wang, Sarna, Tian, Isola, Maschinot, Liu, and Krishnan]{supcon}
Prannay Khosla, Piotr Teterwak, Chen Wang, Aaron Sarna, Yonglong Tian, Phillip Isola, Aaron Maschinot, Ce~Liu, and Dilip Krishnan.
\newblock Supervised contrastive learning.
\newblock \emph{Advances in neural information processing systems}, 33:\penalty0 18661--18673, 2020.

\bibitem[Kim et~al.(2022)Kim, Jain, and Liu]{adaface}
Minchul Kim, Anil~K Jain, and Xiaoming Liu.
\newblock Adaface: Quality adaptive margin for face recognition.
\newblock In \emph{Proceedings of the IEEE/CVF conference on computer vision and pattern recognition}, pages 18750--18759, 2022.

\bibitem[Kim et~al.(2023)Kim, Jeong, and Kwak]{kim2023hier}
Sungyeon Kim, Boseung Jeong, and Suha Kwak.
\newblock Hier: Metric learning beyond class labels via hierarchical regularization.
\newblock In \emph{Proceedings of the IEEE/CVF conference on computer vision and pattern recognition}, pages 19903--19912, 2023.

\bibitem[Liao et~al.(2012)Liao, Jain, and Li]{partialface}
Shengcai Liao, Anil~K Jain, and Stan~Z Li.
\newblock Partial face recognition: Alignment-free approach.
\newblock \emph{IEEE Transactions on pattern analysis and machine intelligence}, 35\penalty0 (5):\penalty0 1193--1205, 2012.

\bibitem[Lin et~al.(2014)Lin, Maire, Belongie, Hays, Perona, Ramanan, Doll{\'a}r, and Zitnick]{lin2014microsoft}
Tsung-Yi Lin, Michael Maire, Serge Belongie, James Hays, Pietro Perona, Deva Ramanan, Piotr Doll{\'a}r, and C~Lawrence Zitnick.
\newblock Microsoft coco: Common objects in context.
\newblock In \emph{European conference on computer vision}, pages 740--755. Springer, 2014.

\bibitem[Movshovitz-Attias et~al.(2017)Movshovitz-Attias, Toshev, Leung, Ioffe, and Singh]{proxynca}
Yair Movshovitz-Attias, Alexander Toshev, Thomas~K Leung, Sergey Ioffe, and Saurabh Singh.
\newblock No fuss distance metric learning using proxies.
\newblock In \emph{Proceedings of the IEEE international conference on computer vision}, pages 360--368, 2017.

\bibitem[Qian et~al.(2020)Qian, Wang, Zhang, Zhu, Fu, Xiang, Jiang, and Xue]{qian2020long}
Xuelin Qian, Wenxuan Wang, Li~Zhang, Fangrui Zhu, Yanwei Fu, Tao Xiang, Yu-Gang Jiang, and Xiangyang Xue.
\newblock Long-term cloth-changing person re-identification.
\newblock In \emph{Proceedings of the Asian conference on computer vision}, 2020.

\bibitem[Ryoo et~al.(2021)Ryoo, Piergiovanni, Arnab, Dehghani, and Angelova]{ryoo2021tokenlearner}
Michael~S Ryoo, AJ~Piergiovanni, Anurag Arnab, Mostafa Dehghani, and Anelia Angelova.
\newblock Tokenlearner: What can 8 learned tokens do for images and videos?
\newblock \emph{arXiv preprint arXiv:2106.11297}, 2021.

\bibitem[Schroff et~al.(2015)Schroff, Kalenichenko, and Philbin]{facenet}
Florian Schroff, Dmitry Kalenichenko, and James Philbin.
\newblock Facenet: A unified embedding for face recognition and clustering.
\newblock In \emph{Proceedings of the IEEE conference on computer vision and pattern recognition}, pages 815--823, 2015.

\bibitem[Siarohin et~al.(2019)Siarohin, Lathuili{\`e}re, Tulyakov, Ricci, and Sebe]{siarohin2019first}
Aliaksandr Siarohin, St{\'e}phane Lathuili{\`e}re, Sergey Tulyakov, Elisa Ricci, and Nicu Sebe.
\newblock First order motion model for image animation.
\newblock \emph{Advances in neural information processing systems}, 32, 2019.

\bibitem[Song et~al.(2022)Song, Wu, Qian, He, and Loy]{song2022everybody}
Linsen Song, Wayne Wu, Chen Qian, Ran He, and Chen~Change Loy.
\newblock Everybody’s talkin’: Let me talk as you want.
\newblock \emph{IEEE Transactions on Information Forensics and Security}, 17:\penalty0 585--598, 2022.

\bibitem[Sun et~al.(2019)Sun, Zheng, Li, Yang, Tian, and Wang]{pcb}
Yifan Sun, Liang Zheng, Yali Li, Yi~Yang, Qi~Tian, and Shengjin Wang.
\newblock Learning part-based convolutional features for person re-identification.
\newblock \emph{IEEE transactions on pattern analysis and machine intelligence}, 43\penalty0 (3):\penalty0 902--917, 2019.

\bibitem[Tran et~al.(2017)Tran, Yin, and Liu]{drgan}
Luan Tran, Xi~Yin, and Xiaoming Liu.
\newblock Disentangled representation learning gan for pose-invariant face recognition.
\newblock In \emph{Proceedings of the IEEE conference on computer vision and pattern recognition}, pages 1415--1424, 2017.

\bibitem[Truong and Venkatesh(2007)]{truong2007video}
Ba~Tu Truong and Svetha Venkatesh.
\newblock Video abstraction: A systematic review and classification.
\newblock \emph{ACM transactions on multimedia computing, communications, and applications (TOMM)}, 3\penalty0 (1):\penalty0 3--es, 2007.

\bibitem[Wan and Chen(2017)]{occlusionface}
Weitao Wan and Jiansheng Chen.
\newblock Occlusion robust face recognition based on mask learning.
\newblock In \emph{2017 IEEE international conference on image processing (ICIP)}, pages 3795--3799. IEEE, 2017.

\bibitem[Wang et~al.(2018{\natexlab{a}})Wang, Yuan, Chen, Li, and Zhou]{mgn}
Guanshuo Wang, Yufeng Yuan, Xiong Chen, Jiwei Li, and Xi~Zhou.
\newblock Learning discriminative features with multiple granularities for person re-identification.
\newblock In \emph{Proceedings of the 26th ACM international conference on Multimedia}, pages 274--282, 2018{\natexlab{a}}.

\bibitem[Wang et~al.(2018{\natexlab{b}})Wang, Wang, Zhou, Ji, Gong, Zhou, Li, and Liu]{cosface}
Hao Wang, Yitong Wang, Zheng Zhou, Xing Ji, Dihong Gong, Jingchao Zhou, Zhifeng Li, and Wei Liu.
\newblock Cosface: Large margin cosine loss for deep face recognition.
\newblock In \emph{Proceedings of the IEEE conference on computer vision and pattern recognition}, pages 5265--5274, 2018{\natexlab{b}}.

\bibitem[Wang et~al.(2018{\natexlab{c}})Wang, Liu, Zhu, Liu, Tao, Kautz, and Catanzaro]{wang2018video}
Ting-Chun Wang, Ming-Yu Liu, Jun-Yan Zhu, Guilin Liu, Andrew Tao, Jan Kautz, and Bryan Catanzaro.
\newblock Video-to-video synthesis.
\newblock \emph{arXiv preprint arXiv:1808.06601}, 2018{\natexlab{c}}.

\bibitem[Xu et~al.(2025)Xu, Guo, Hu, Chu, Wang, He, Wang, Shi, He, Zhu, et~al.]{xu2025qwen3}
Jin Xu, Zhifang Guo, Hangrui Hu, Yunfei Chu, Xiong Wang, Jinzheng He, Yuxuan Wang, Xian Shi, Ting He, Xinfa Zhu, et~al.
\newblock Qwen3-omni technical report.
\newblock \emph{arXiv preprint arXiv:2509.17765}, 2025.

\end{thebibliography}

\end{document}